\documentclass{article} 
\usepackage[accepted]{icml2016}
\usepackage{hyperref}
\usepackage{url}
\usepackage{graphicx}
\usepackage{amsmath, amssymb}
\usepackage{subfig}
\usepackage{todonotes}
\usepackage{xspace} 
\usepackage{multirow}
\usepackage{multicol}
\usepackage{colortbl}

\definecolor{TUMBlau}{RGB}{0,101,189}

\let\originalleft\left
\let\originalright\right
\renewcommand{\left}{\mathopen{}\mathclose\bgroup\originalleft}
\renewcommand{\right}{\aftergroup\egroup\originalright}

\newcommand{\EE}{{\mathbb{E}}}

\newcommand{\bx}{\ensuremath{\mathbf{x}}}
\newcommand{\xt}{\bx_{t}}
\newcommand{\xtprev}{\bx_{t-1}}

\newcommand{\xTs}{\bx_{1:T}}

\newcommand{\bz}{\ensuremath{\mathbf{z}}}
\newcommand{\zt}{\bz_{t}}
\newcommand{\ztprev}{\bz_{t-1}}

\newcommand{\zTs}{\bz_{1:T}}

\newcommand{\bh}{\ensuremath{\mathbf{h}}}
\newcommand{\htemp}{\bh_{t}}
\newcommand{\htprev}{\bh_{t-1}}

\newcommand{\hTs}{\bh_{0:T}}

\newcommand{\p}[1]{p\left(#1\right)}
\newcommand{\pgiven}[2]{{\ensuremath{\p{#1\mid#2}}}}
\newcommand{\pundergiven}[3]{{\ensuremath{p_{#2}\left(#1\mid#3\right)}}}

\newcommand{\qundergiven}[3]{{\ensuremath{q_{#2}\left(#1\mid#3\right)}}}

\newcommand{\expectunder}[2]{{\ensuremath{\EE_{#2}\left[#1\right]}}}

\newcommand{\KL}[2]{\ensuremath{\operatorname{KL}\left(#1\mid\mid#2\right)}}

\newcommand{\sect}[1]{Section~#1}

\newcommand{\fig}[1]{Fig.~#1}

\newcommand{\qphi}{q_\phi}

\icmltitlerunning{Variational Inference for On-line Anomaly Detection in High-Dimensional Time Series}

\begin{document} 
	
	\twocolumn[
	\icmltitle{Variational Inference for On-line Anomaly Detection in High-Dimensional Time Series}
	
	\icmlauthor{Maximilian Soelch \& Justin Bayer}{zip([m.soelch, bayer], [@tum.de, @sensed.io])}
	\icmladdress{Chair of Robotics and Embedded Systems\\Department of Informatics,	Technische Universit\"at M\"unchen, Germany}
	\icmlauthor{Marvin Ludersdorfer \& Patrick van der Smagt}{zip([ludersdorfer], [@fortiss.org])}
	\icmladdress{fortiss, An-Institut Technische Universit\"at M\"unchen, Germany}
	
	\icmlkeywords{boring formatting information, machine learning, ICML}
	
	\vskip 0.3in
	]

\begin{abstract}
Approximate variational inference has shown to be a powerful tool for modeling unknown complex probability distributions. Recent advances in the field allow us to learn probabilistic models of sequences that actively exploit spatial and temporal structure. We apply a Stochastic Recurrent Network (STORN) to learn robot time series data. Our evaluation demonstrates that we can robustly detect anomalies both off- and on-line.
\end{abstract}

\section{Introduction}
With a complex system like a robot, we would like to be able to discriminate between normal and anomalous behavior of this system. For instance, we would like to be able to recognize that something went wrong while the robot was fulfilling a task. Generally speaking, determining whether an unknown sample is structurally different from prior knowledge is referred to as anomaly detection.

Recording anomalous data is costly (or even dangerous) in comparison to normal data. Moreover, anomalies are inherently diverse, which prohibits explicit modeling. Due to the underrepresentation of anomalous samples in training data, anomaly detection remains a challenging instance of two-class classification to this day. Consequently, the problem is reversed: A normality score is learned from normal data only, and the fully trained normality score is used to discriminate anomalous from normal data by thresholding.

The contribution of this paper is an application of approximate variational inference for anomaly detection. We learn a generative time series model of the data, which can handle high-dimensional, spatially and temporally structured data. Neither the learning algorithm nor subsequent anomaly detection via scores requires domain knowledge.

\section{Problem Description: Anomaly Detection}
As \cite{anomalyreview2014} show, a plethora of anomaly detection approaches exist. A common assumption for time series anomaly detection is that data streams are i.i.d.\ in time and/or space. For robots (and many other systems) this is not true: Joint torques that are perfectly normal in one joint configuration are anomalous in another. This is not captured by previous approaches. A notable exception is \cite{milacskirobust}. However, their approach, requiring the entire time series for processing, lacks on-line capability and their off-line evaluation cannot be transferred to different models because it is based on a model-based segmentation of the time-series. \cite{an2015variational} have independently applied variational inference anomaly detection on static data. Since no comparable algorithm exists, no comparison is possible.

\begin{figure}
	\includegraphics[width=\columnwidth]{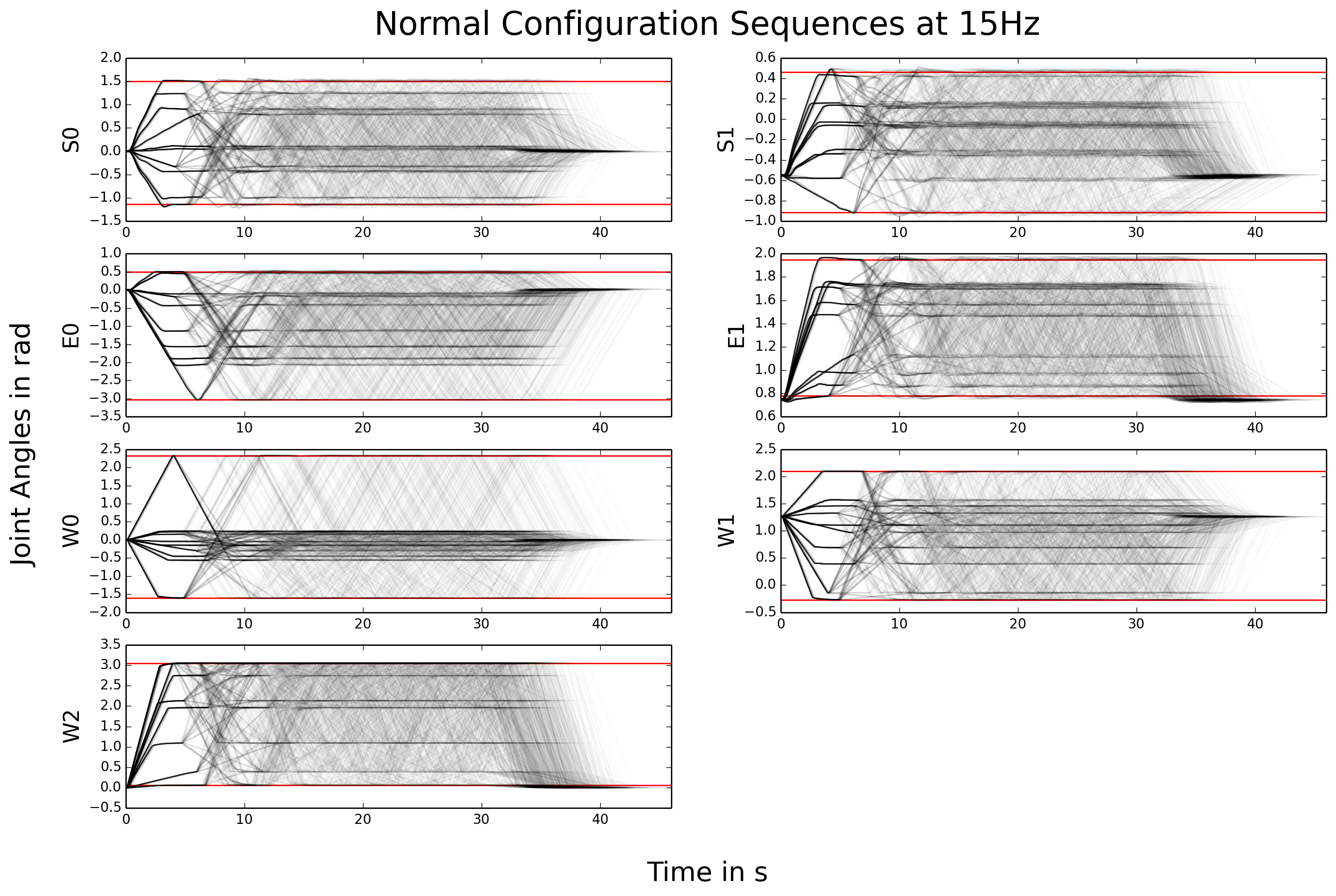}
	\caption{1008 normal samples. This approximates the target distribution we are trying to learn with STORN.}\label{fig:data}
\end{figure}

For training and testing, we recorded the joint configurations of the seven joints of a Rethink Robotics Baxter Robot arm. We recorded 1008 anomaly-free samples at 15 Hz of a pick-and-place task, our target distribution, cf.\ \fig{\ref{fig:data}}. This task is simulated by first fixing a pool of 10 waypoints and then, for each sample, traversing a random sequence of as many waypoints as possible for a duration of 30s and finally returning to the initial configuration. This results in roughly 8 to 10 waypoints per sample. For this distribution, we would like to learn a generative model.

For testing purposes, we recorded 300 samples with anomalies obtained by manually hitting the robot on random hit commands. For each time stamp, we obtained two labels: whether or not a hit command occurred within the previous 4 seconds (a rough bound on the human response time), and unusual torque.\footnote{Torque was only used for labeling, not for learning.} Both are depicted as red and blue background color in \fig{\ref{fig:online}}. Neither of these labels is perfect, the temporal label is necessarily too loose, whereas the torque label misses subtle anomalies while putting false positive labels on artifacts in the data.

\begin{figure}
	\centering
	\includegraphics[width=\columnwidth]{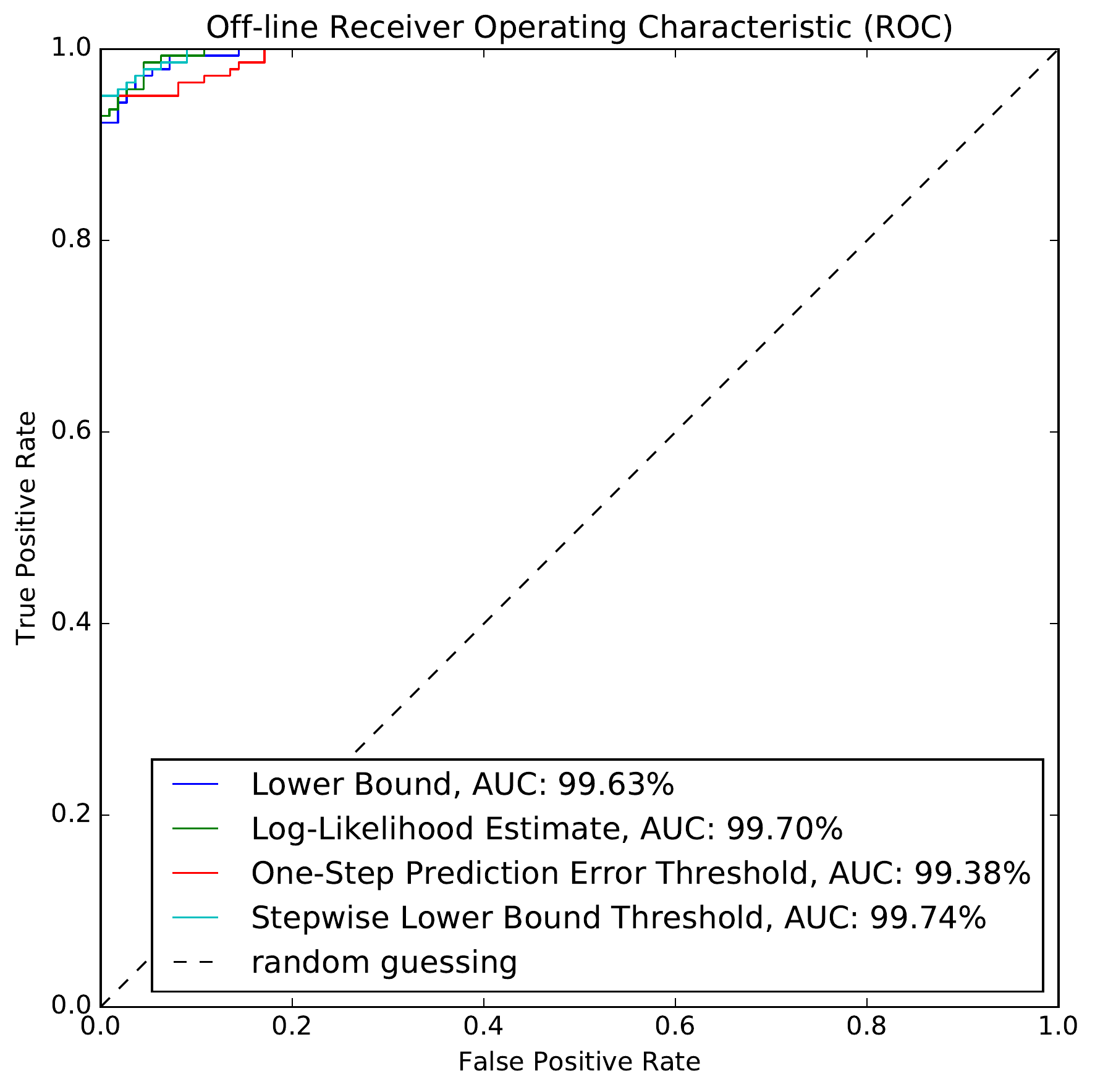}
	\caption{Receiver operating characteristic (ROC) for evaluating the off-line detection algorithm for different normality criteria and thresholds.}
	\label{fig:offline}	
\end{figure}
\begin{table*}[t]	\caption{Standard metrics for classification on \emph{unseen} test set data.  PPV and NPV denote positive and negative predictive value, respectively. Each cell contains four values, from left to right the result for the four scores in the order outlined in \sect{\ref{sub:off}}.}\centering
	\begin{tabular}{llrrrrrrrrrrrrr}
		\hline
		&&\multicolumn{8}{c}{Anomaly}&\\
		&&\multicolumn{4}{l}{Positive}&\multicolumn{4}{l}{Negative}&&&&\\\hline
		\multirow{2}{*}{Detect.}&P&
		\cellcolor{TUMBlau!100}141&\cellcolor{TUMBlau!90}141&\cellcolor{TUMBlau!80}142&\cellcolor{TUMBlau!70}140&
		\cellcolor{TUMBlau!40}0&\cellcolor{TUMBlau!30}3&\cellcolor{TUMBlau!20}2&\cellcolor{TUMBlau!10}1&
		1.0&.979&.986&.993&\emph{PPV}\\
		&N&
		\cellcolor{TUMBlau!40}4&\cellcolor{TUMBlau!30}4&\cellcolor{TUMBlau!20}3&\cellcolor{TUMBlau!10}5&
		\cellcolor{TUMBlau!100}109&\cellcolor{TUMBlau!90}106&\cellcolor{TUMBlau!80}107&\cellcolor{TUMBlau!70}108&
		.965&.964&.973&.956&\emph{NPV}\\\hline
		&&.972&.972&.979&.966&1.0&.972&.982&.991&.984&.972&.980&.976&\\
		&&\multicolumn{4}{l}{\emph{Sensitivity}}&\multicolumn{4}{l}{\emph{Specificity}}&\multicolumn{4}{l}{\emph{Accuracy}}\\\hline
	\end{tabular}
	\label{tab:metrics}
\end{table*}
\section{Methodology: Variational Inference and Stochastic Recurrent Networks}
In the wake of \cite{dlgm2014, vae2013}, who introduced the Variational Auto-Encoder (VAE), there has been a renewed interest in variational inference. The VAE replaces an auto-encoder's latent representation $\bz$ of given data $\bx$ with stochastic variables.

The decoding distribution (or \emph{recognition model}) $\qundergiven{\bz}{\phi}{\bx}$ approximates the true, \emph{unknown} posterior $\pundergiven{\bz}{}{\bx}$. The encoding distribution $\pundergiven{\bx}{\theta}{\bz}$ implements a simple graphical model. Decoder and encoder are parametrized by $\phi$ and $\theta$, respectively, e.g., neural networks. This allows to overcome intractable posterior distributions by learning.

The approach is theoretically justified by the observation that an arbitrary inverse transform sampling is approximated.

The approach is generalized to time series by \cite{storn2014} by applying \emph{recurrent} neural networks with hidden layers as encoder and decoder.  For observations $\xTs$ and corresponding latent states $\zTs$, we assume the factorization 
\begin{align}
\p{\xTs, \zTs} = \prod_{t=1}^{T} &\pundergiven{\xt}\theta{\htemp^{g} (\xtprev,\zt,\htprev^{g})     }\\&\cdot\pundergiven{\bz_t}\psi{\htemp^{p}(\ztprev, \htprev^{p})}\label{eq:nfactorization}
\end{align}
The two distributions, the generative model (decoder) and the prior over latents, are implemented by two RNNs with parameters $\theta$ and $\psi$ and hidden states $\htemp^g$ and $\htemp^p$, respectively. These RNNs output sufficient statistics of normal distributions. This introduces a \emph{trending prior}, i.e., a prior that can change over time, which is an extension of the original STORN. Also, it forms an extension of \cite{vrnn2015}.

With the encoding RNN  with parameters $\phi$ implementing the recognition model $\qundergiven{\zTs}{\phi}{\xTs}$, we arrive at the typical variational lower bound (also referred to as free energy) to the marginal log-likelihood $\ln\p{\xTs}$:
\begin{align}
\mathcal{L}\left(\qphi\right)&:=\,\expectunder{\sum_{t=1}^{T}\ln \pgiven{\xt}{\htemp^{g}}}{\qphi} \label{eq:loss}\\\nonumber
&\hphantom{\mathbb{E}_{\qphi}}-\KL{\qundergiven{\zTs}{\phi}{\xTs}}{\pgiven{\zTs}{\hTs^{p}}}\\
&\leq\ln\p{\xTs}\label{eq:ll}
\end{align}
Maximization of the lower bound is provably equivalent to minimizing the KL-divergence between the true posterior $\pgiven{\bz}{\bx}$ and the approximate posterior $\qundergiven{\bz}{\phi}{\bx}$.
The lower bound can be used to simultaneously train all adjustable parameters by stochastic backpropagation. The decomposition over time allows for computationally feasible on-line detection.

For each time step, STORN outputs a lower bound value and a
predictive distribution for the next time step. These (and post-processed
versions) serve as scores for thresholding anomalies.

\section{Experiments}
Prior to any anomaly detection, we trained STORN on a training set of 640 normal time series. Model selection was then based on 160 validation samples.
Based on a fixed STORN, the anomaly detection then derives a scalar score from the outputs of the model and finds a threshold to discriminate normal from anomalous data. Anomaly detection was tested on the 208 remaining normal samples and 300 anomalous samples: Half of these 508 samples were taken to extract thresholds and the remaining half was taken to test overall performance of the detection algorithm.
\subsection{Off-line Detection}\label{sub:off}

For off-line detection, i.e., detecting whether an unknown test sample has an anomaly or not, we used different normality scores:
\begin{enumerate}
	\item Lower bound (value of objective function (\ref{eq:loss})).
	\item Monte Carlo estimate of the true likelihood (\ref{eq:ll}) (for comparative reasons).
	\item A high percentile (outlier-adjusted extreme value) of the deviation from MAP one-step prediction.
	\item A high percentile of the step-wise components of the lower bound.
\end{enumerate}

It should be stressed that none of the normality scores is related to the original data domain. Anomaly detection is entirely transferred onto probabilistic grounds.

A ROC curve on the 254 samples used for threshold extraction can be seen in \fig{\ref{fig:offline}}. For each of the four scores, we used the threshold minimizing the sum of squared sensitivity and specificity (which coincides with the threshold corresponding to the point closest to the top left corner of the ROC curve---the perfect classifier).

Table~\ref{tab:metrics} reports standard metrics for classification on unseen test data. We see that off-line classification is remarkably robust.

\subsection{On-line Detection}
\begin{figure}
	\centering
	\includegraphics[width=\columnwidth]{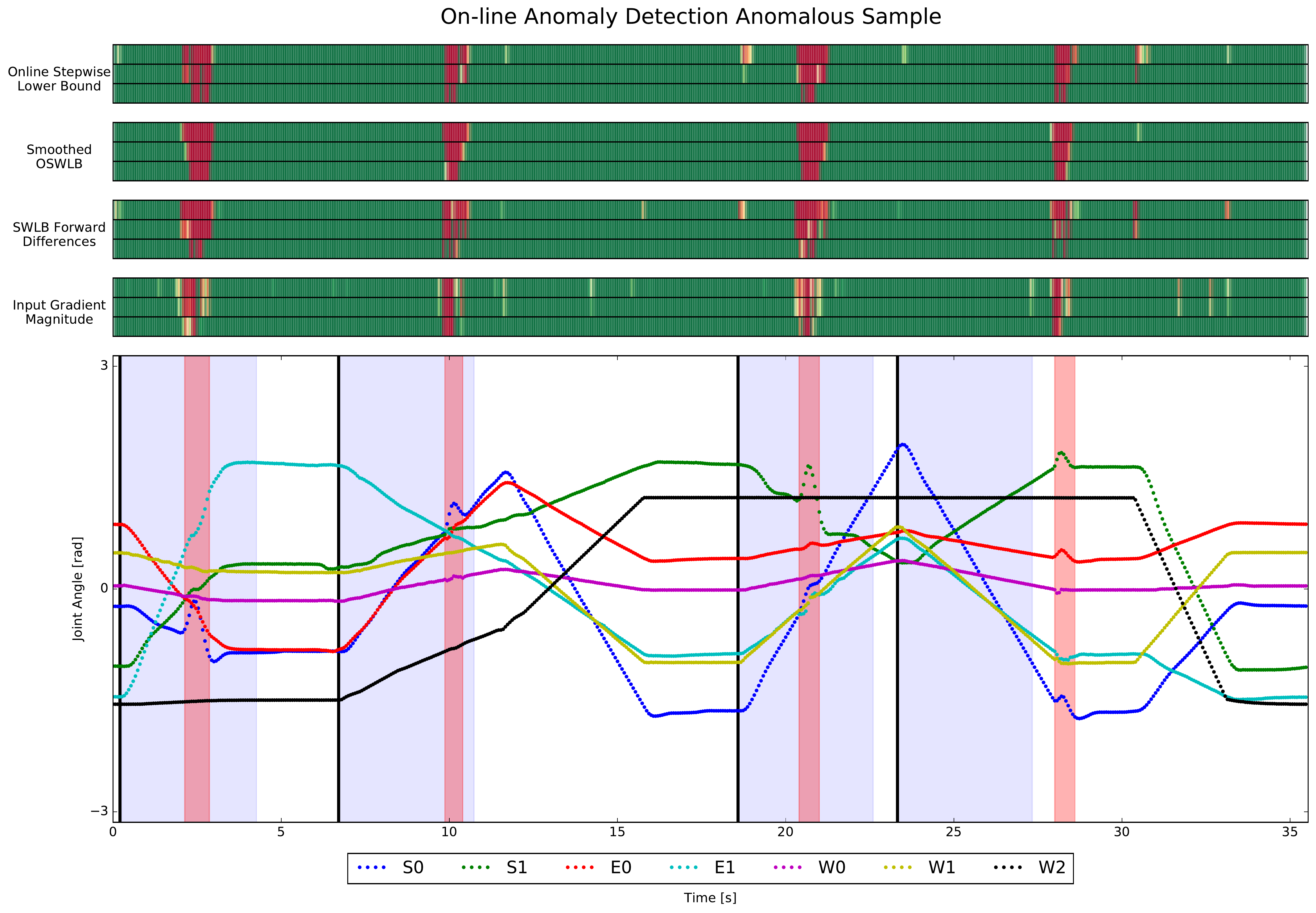}
	\caption{Anomalous test sample with labels (red and blue background colors), hit commands (black lines). On-line normality scores color-coded by the 12 bars (4 scores, 3 thresholds each)---green-red-yellow spectrum indicating that the respective score is below, on, or above the anomaly threshold.}
	\label{fig:online}
\end{figure}

The more challenging case of on-line detection is depicted in \fig{\ref{fig:online}}. Again, we applied four different normality scores. Three were based on the step-wise lower bound---we used the step-wise lower bound output of our model, as well as a smoothed version (with a narrow Gaussian Kernel), and the absolute value of forward differences. As a fourth score, we used the step-wise magnitude of lower-bound gradient w.r.t.\ the sample. A large gradient magnitude in one time step indicates a significant perturbation from a more likely time series; such perturbations are indicators of anomalous data.
As with the off-line scores, none of the on-line scores is related to the original data domain. This renders the overall approach very flexible.

For each of the four scores, we extracted three different thresholds, each leveraging the two types of imperfect labels differently. For each of the four scores, we extracted three thresholds, namely the ones maximizing
\begin{enumerate}
	\item the sum of squared sensitivity and specificity on torque-based labels,
	\item in addition to 1.\ the positive predictive value (PPV) on torque labels,
	\item in addition to 1.\ and 2.\ the weighted PPV on hit-command-based labels.
\end{enumerate}
The underlying assumption is that we do not necessarily want to perfectly recover the true labels (which we do not know), but we want to spot anomalies qualitatively (i.e., report an anomaly while it is ongoing while at the same time having few false alarms). The choice of metric for extracting a threshold highly depends on the application at hand.

These four times three thresholds can be seen in \fig{\ref{fig:online}}. 

\section{Conclusion and Future Work}
In this paper, we successfully applied the framework of variational inference (VI), in particular Stochastic Recurrent Networks (STORNs), for learning a probabilistic generative model of high-dimensional robot time series data. No comparable approach has been proposed previously.

This new approach enables feasible off- and on-line detection without further assumptions on the data. In particular, no domain knowledge is required for applying both the learning and the detection algorithm. This renders our algorithm a very flexible, generic approach for anomaly detection in spatially and temporally structured time series. We have shown that the new approach is able to detect anomalies in robot time series data with remarkably high precision.

Future research will have to show reproducibility of the results (i) with different kinds of anomalies, (ii) in new environments (e.g., on other robots). 

Furthermore, we believe that variational inference will enable us to extract the true latent dynamics of the system from observable data by introducing suitable priors and transitions into STORN. This will equip us with a more meaningful latent space, which can in turn serve as a basis for new detection methods based on output of STORN.
\subsubsection*{Acknowledgments}
A previous version of this paper was presented at the Workshop of ICLR 2016.

Part of this work has been supported by the
TAC-MAN project, EC Grant agreement no. 610967,
within the FP7 framework program.

{Patrick van der Smagt is also affiliated with BRML, Technische Universit\"at M\"unchen, Germany.}

\bibliography{Bibliography}
\bibliographystyle{icml2016}
\end{document}